\def\year{2019}\relax
\documentclass[letterpaper]{article} %DO NOT CHANGE THIS
\usepackage{aaai19}  %Required
\usepackage{times}  %Required
\usepackage{helvet}  %Required
\usepackage{courier}  %Required
\usepackage{url}  %Required
\usepackage{graphicx}  %Required
\usepackage[pdftex]{hyperref}
\usepackage{amsmath}
\usepackage{color}
\usepackage{textcomp}
\usepackage{booktabs}
\usepackage{ccicons}
\usepackage{todonotes}
\usepackage{amsfonts}
\usepackage{framed}
\usepackage{enumitem}
\usepackage{subcaption}

\usepackage{tikz}
\usetikzlibrary{shapes}

\usepackage[normalem]{ulem}
\usepackage{tikz}
\usetikzlibrary{arrows}
\usepackage{framed}
\usepackage{tabularx}
\usepackage{listings}

\tikzset{
	main node/.style={circle,draw,font=\small},
	rectangle/.style={font=\small}
}

\makeatletter
  \let\@internalcite\cite
  \def\cite{\def\citeauthoryear##1##2{##1, ##2}\@internalcite}
  \def\shortcite{\def\citeauthoryear##1{##2}\@internalcite}
  \def\@biblabel#1{\def\citeauthoryear##1##2{##1, ##2}[#1]\hfill}
\makeatother

\newcommand{\cA}{\mathcal{A}}

\newcommand{\tit}[1]{\textit{#1}}

\newcommand{\CD}{\mathit{CD}}

\newcommand{\figref}[1]{Figure~\ref{fig:#1}}

\begin{document}
\title{Human-In-The-Loop Learning of Qualitative Preference Models}
\author{
	Joseph Allen\\
	School of Computing\\
	University of North Florida\\
	Jacksonville, FL\\
	n01045721@unf.edu
	\And
	Ahmed Moussa\\
	School of Computing\\
	University of North Florida\\
	Jacksonville, FL\\
	nagar@aucegypt.edu
	\And
	Xudong Liu \\
	School of Computing\\
	University of North Florida\\
	Jacksonville, FL\\
	xudong.liu@unf.edu
}
\maketitle

\begin{abstract}
In this work, we present a novel human-in-the-loop framework to help the human user
understand the decision making process that involves choosing preferred options.
We focus on qualitative preference models over alternatives from combinatorial domains.
This framework is \tit{interactive}:
the user provides her behavioral data to the framework, 
and the framework explains the learned model to the user.
It is \tit{iterative}:
the framework collects feedback on the learned model from the user and tries to
improve it accordingly till the user terminates the iteration.
In order to communicate the learned preference model to the user, we develop
visualization of intuitive and \tit{explainable} graphic 
models, such as lexicographic preference trees and forests, and conditional
preference networks.
To this end, we discuss key aspects of our framework for lexicographic 
preference models.
\end{abstract}

\section{Introduction}
%\begin{enumerate}[itemsep=0pt]
%  \item What is the problem we are solving in this paper?
%  \begin{itemize}[itemsep=0pt]
%    \item Human-in-the-loop learning of lexicographic preference models.
%  \end{itemize}
%  \item Why is it important and challenging to solve?
%  \begin{itemize}[itemsep=0pt]
%    \item Learning explainable decision models is very important. \cite{gunning2017explainable}
%  \end{itemize}
%  \item What are the results in the literature already?
%  \begin{itemize}[itemsep=0pt]
%    \item Learning preference trees and forests. \cite{booth:learningLP,conf/aaai15/LiuT,conf/foiks18/LiuT}
%    \item Learning conditional preference networks. \cite{bbdh03,allen2017learning}
%  \end{itemize}
%  \item What are our approach and results?
%\end{enumerate}

Preferences are an essential component to decision making and have 
been extensively studied in research communities such as
decision theory, computational social choice, recommender systems, and
knowledge representation.
Various preference models have been proposed in the literature to
represent preferences of different types including two major ones:
quantitative models and qualitative models.
Quantitative preference models integrate into the models numeric 
values used to define the preference relation of objects.
These models include fuzzy constraint satisfaction 
models \cite{schiex1992possibilistic}, 
penalty logic \cite{de1994penalty}, and
possibilistic logic \cite{dubois1994possibilistic}.
On the other hand, qualitative preference models describe,
either directly or indirectly, the relative ordinal relation of objects.
Such models include
lexicographic preference trees (LP-trees) \cite{booth:learningLP,conf/adt13/LiuT,conf/adt15/LiuT},
lexicographic preference forests (LP-forests) \cite{conf/gcai16/liuT,conf/foiks18/LiuT},
conditional preference network (CP-nets) \cite{bbdh03}, and
answer set optimization \cite{brewka2003answer}.
These models draw our focus because they are proven to be intuitive,
cognitively plausible, and predictive with high 
accuracy \cite{allen2015beyond,conf/foiks18/LiuT}.

In this paper, we focus on the learning problem of qualitative preference 
models, in particular, graphical models that are intuitive and often
compact in size, such as LP-trees, LP-forests and CP-nets.
Recently, active and passive learning of these graphical models have been 
studied, both theoretically and empirically, in the 
community \cite{conf/aaai15/LiuT,conf/foiks18/LiuT,koriche2010learning,alanazi2016complexity,allen2017learning}.
However, these traditional preference learning works do not 
leverage the intuitivity and explainability of the models to
\tit{interact} with the decision maker in the learning process.
Models explainable to human users are desirable when decision makers
in various applications are to understand or even trust the resulting models
formulated by intelligent machine partners \cite{gunning2017explainable}.

To this end, we propose a novel framework that learns qualitative
preference models.
This framework is interactive and iterative:
from the decision making user it obtains behavioral data, which is then
preprocessed before sent to a preference learner that computes
models (i.e., an LP-tree, LP-forest or CP-net) to be visualized and presented
back to the user for feedback in order to improve the models in the following
iterations.
This learning process terminates when the user is satisfied with the learned
models.
In this report, we show our design and implementation of the framework,
so far for learning LP-trees and LP-forests, that is a web application using
Django with Python and C++ as the programming languages on the server.

In the next section, we define and exemplify the two models: LP-trees and LP-forests.
Then, we present our human-in-the-loop preference learning framework,
and demonstrate it by showing our prototype that learns LP models.
Finally, we conclude, pointing to possible future research directions.

\section{Lexcigraphic Preference Trees and Forests}
%\begin{enumerate}[itemsep=0pt]
%  \item What are lexicographic preference trees?
%  \item What are lexicographic preference forests?
%  \item Provide examples.
%\end{enumerate}

The preference models we consider in this work are over alternatives from
combinatorial domains of multi-valued attributes.  
Now we define combinatorial domains and preference models we focus in our
paper including lexicographic preference trees and forests.
Let $\cA=\{X_1,\ldots,X_p\}$ be a set of categorical attributes, each
$X_i$ with a finite domain $D_i$, where $|D_i|$ is bounded by a constant.
The \tit{combinatorial domain} $\CD(\cA)$ over $\cA$ is the Cartesian product 
$D_1 \times \ldots \times D_p$. Elements of combinatorial
domains are called \tit{alternatives}. 

A lexicographic preference tree over $\CD(\cA)$ is an ordered labeled tree, where (1) every
non-leaf node is labeled by an attribute $X_i \in \cA$, and by
a \emph{local preference} $>_i$, which is a total order over
$D_i$; (2) every non-leaf node labeled by an attribute $X_i$ has $|D_i|$
outgoing edges, ordered from left to right according to $>_i$;
(3) every 
leaf node is denoted by $\Box$; and (4) on every path from the root to a 
leaf each attribute appears \emph{at most once} as a label. 
Each tree induces a \tit{total preorder} that precisely is defined by
the order of the leafs.

\begin{figure}[!ht]
	\centering
  \scalebox{.95}{
	\begin{tikzpicture}[->,>=stealth',
	  level 1/.style={sibling distance=2.5cm, level distance=35pt},
		level 2/.style={sibling distance=0.7cm, level distance=30pt},
		level 3/.style={sibling distance=0.4cm, level distance=30pt}]
	  \node [main node,inner sep=3pt,label={[xshift=1.1cm, yshift=-0.5cm]$v>s>r$}] {$B$}
	    child {node [main node,inner sep=3pt,label={[xshift=1.1cm, yshift=-0.6cm]$ d>l>g$}] {$P$}
	  		child {node [rectangle,draw,label=below:0] {}}
	  		child {node [rectangle,draw,label=below:1] {}}
	  		child {node [rectangle,draw,label=below:2] {}}
	    }
	    child {node [main node,inner sep=2.5pt,label={[xshift=0.8cm, yshift=-0.6cm]$ h>f$}] {$M$}
	  		child {node [rectangle,draw,label=below:3] {}}
	  		child {node [rectangle,draw,label=below:4] {}}
			}
	  	child {node [rectangle,draw,label=below:5] {}}
	    %child {node [main node,inner sep=3.3pt,label={[xshift=0.9cm, yshift=-0.6cm]$ m>a$}] {$T$}
	  	%	child {node [rectangle,draw] {}}
	  	%	child {node [rectangle,draw] {}}
	    %}
		;
	\end{tikzpicture}
	}
  \caption{A preference tree $T$ over the car domain\label{fig:LPT_full_ex}}
\end{figure}
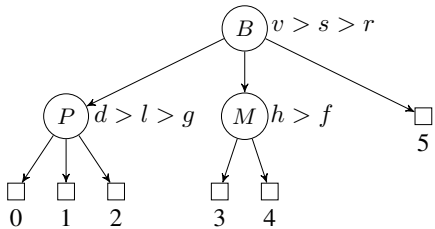

To illustrate, let us consider the domain of cars described by four 
attributes: \tit{BodyType} ($B$) with 
values: \tit{minivan} ($v$), \tit{sedan} ($s$), and \tit{sport} ($r$);
\tit{Make} ($M$) with values \tit{Honda} ($h$) 
and \tit{Ford} ($f$); \tit{Price} ($P$) with 
values \tit{low} ($l$), \tit{medium} ($d$), and \tit{high} ($g$); and
\tit{Transmission} ($T$) with \tit{automatic} ($a$) and \tit{manual} ($m$). 
An user's preference
order on cars from this space could be expressed by a tree $T$ in 
\figref{LPT_full_ex}. 

Tree $T$ informs us that the most important attribute is \tit{BodyType} with
the user preferring minivans the most, then sedans and sports the least.
Among minivans, the most important attribute is $Price$ with medium preferred
to low to high. Other non-leaf nodes in the tree are interpreted similarly.
Leaf nodes, however, represent sets of cars with the instantiations of the
attributes along their paths.

Given an alternative $o$, we can traverse the tree and find its leaf.
To compare two alternatives, we say that they are \tit{equivalent}
if they have same leaf.  If they have different leaf nodes, the alternative
in the preceding leaf is the preferred one.
For instance, a Honda sedan is better than a Ford sedan, because the former
ends up in leaf 3, preceding leaf 4, the leaf the latter car has.

\subsection{Types of LP-Trees}
\begin{figure}[!ht]
	\centering
  \begin{subfigure}[t]{0.22\textwidth}
		\centering
    \scalebox{.95}{
	    \begin{tikzpicture}[->,>=stealth',node distance=40pt]
	      \node[main node] (1) {$B$};
	      \node[rectangle,draw] at (1.2,0) {$s>v>r$};
	
	      \node[main node] (2) [below of=1] {$M$};
	      \node[rectangle,draw] at (1,-1.4) {$f>h$};

	      \node[main node] (3) [below of=2] {$P$};
	      \node[rectangle,draw] at (1.2,-2.8) {$d>g>l$};
	
		    \path[every node/.style={font=\small}]
	        (1) edge (2)
	        (2) edge (3);
	    \end{tikzpicture}
    }
  	\caption{UIUP\label{fig:trees1}}
	\end{subfigure} \quad
  \begin{subfigure}[t]{0.22\textwidth}
		\centering
    \scalebox{.95}{
	    \begin{tikzpicture}[->,>=stealth',node distance=40pt]
	      \node[main node] (1) {$B$};
	      \node[rectangle,draw] at (1.2,0) {$s>v>r$};
	
	      \node[main node] (2) [below of=1] {$M$};
	      \node[rectangle split, rectangle split parts=3, draw, font=\sffamily\small] at (1.2,-1.3)
		        {
		          $s:h>f$
		          \nodepart{second}
		          $v:f>h$
		          \nodepart{third}
		          $r:h>f$
		        };
	      \node[main node] (3) [below of=2] {$P$};
	      \node[rectangle split, rectangle split parts=2, draw, font=\sffamily\small] at (1.5,-3)
		        {
		          $h:d>l>g$
		          \nodepart{second}
		          $f:l>m>g$
		        };
	
		    \path[every node/.style={font=\small}]
	        (1) edge (2)
	        (2) edge (3);
	    \end{tikzpicture}
    }
  	\caption{UICP\label{fig:trees2}}
	\end{subfigure} 
  \caption{UIUP and UICP LP-trees\label{fig:trees}}
\end{figure}
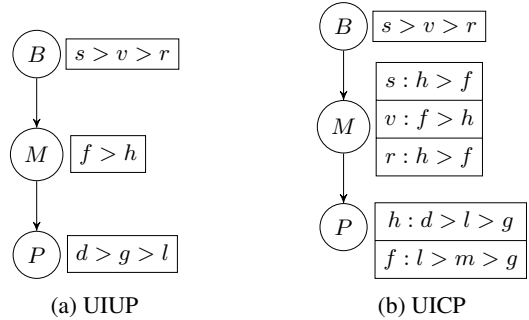

LP-trees, in general, can be of size exponential in the size of the combinatorial
domain.  However, trees with special structures can be collapsed to achieve
compact representation.  When the labeling attributes on all paths of the tree
are exactly the same and the local preference orderings are the same on same
attributes, this tree can be collapsed to a list of nodes labeled by attributes
and unconditional preference orders.
We call this type of LP-trees \tit{unconditional importance and unconditional 
preference} LP-trees (\tit{UIUP LP-trees}).
Keeping this tree structure, if the local preferences are different on same
attributes, these trees can also be collapsed, but to a list of nodes labeled
by attributes and tables of conditional preference orders.
We call this type of LP-trees \tit{unconditional importance and conditional 
preference} LP-trees (\tit{UICP LP-trees}).
All the other LP-trees that are uncollapsible are called 
\tit{conditional importance and conditional preference} 
LP-trees (\tit{CICP LP-trees}), for the importance order of nodes depends on
how their ancestors are instantiated in the tree.
We show examples of these compact representations in \figref{trees1}
and \figref{trees2}.
One example of a CICP tree is the one in \figref{LPT_full_ex}.

\subsection{LP-Forests}
An LP-forest is a finite ensemble of LP-trees over combinatorial domains.
To compare alternatives using an LP-forest, researchers have proposed to 
apply a plethora of voting rules (such as Borda's and Copeland's rules)
to aggregate the decisions of member trees \cite{lang2018voting,conf/foiks18/LiuT}.
We propose to study the visualization problem of big forests of trees
and how to effectively present the forest to the user.
Clearly, visualizing and presenting the whole forest is infeasible.
In this paper, our approach is to only present representative trees in the 
forest according to some distance measure, for which we consider Kendall's
$\tau$ distance.
This measurement calculates the number of pairwise disagreements between
two orderings.  Clearly, it directly applies to computing distances between
LP-trees, for LP-trees represent total orders, after possible equivalent alternatives
are broken alphabetically.
However, if we compute $\tau(T_1,T_2)$ via computing their total orders first,
the process may take time exponential in the size of the two trees if they
are compactly represented of type UIUP or UICP.
To alleviate this, we resort to polynomial algorithms proposed by
Li and Kazimipour \cite{li2018efficient}.
We implemented these algorithms to compute the $\tau$ distances for all
pairs of trees in the forest.
Then, the trees are clustered based on these distance values.

\section{Framework}
\begin{figure}[!ht]
  \centering
    \includegraphics[width=0.5\textwidth]{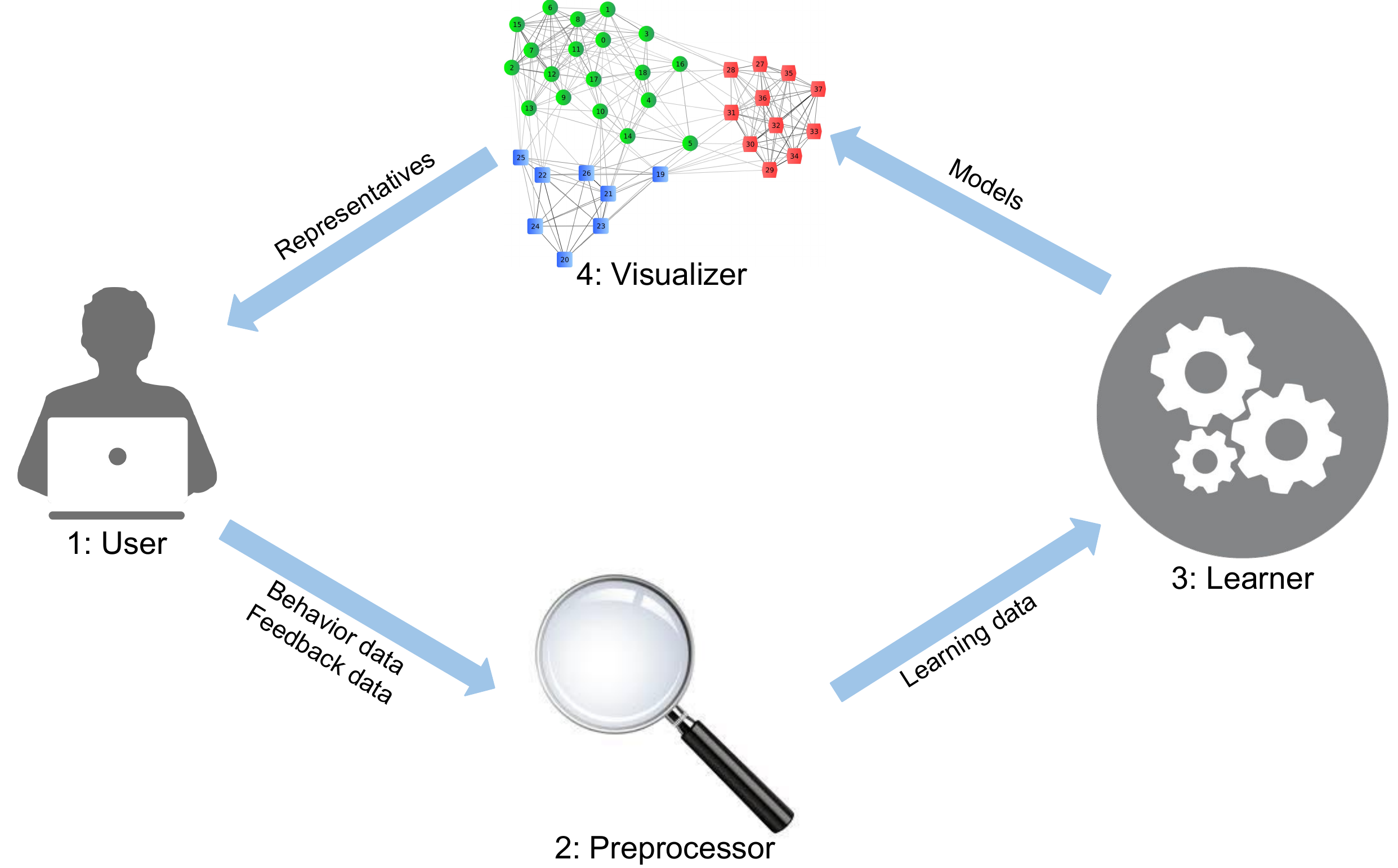}
  \caption{ILPref \label{fig:framework}}
\end{figure}

We now introduce our framework, shown in \figref{framework}, 
for interactive learning of
qualitative preference models in the following.
We call our framework ILPref for short.
The goal of ILPref is to learn, and help the user to understand,
her preferential decision making process over complex domains of options.

\subsection{User}
\begin{figure}[!ht]
  \centering
    \includegraphics[width=0.5\textwidth]{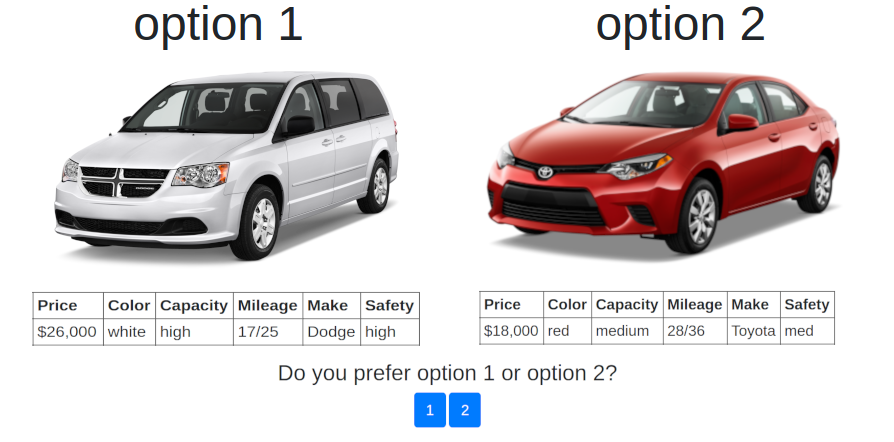}
  \caption{Preferential data collection \label{fig:elicit}}
\end{figure}

The user is the central decision maker, whom the framework tries to
help understand her decision making process.
The user provides \tit{behavioral data} that can be either explicit or implicit.
Explicit data are such as query answers and scaled ratings, whereas
implicit data can be time or clicking distribution over a set of
options reviewed by her.
These are the source data our framework is learning the decision model from.
Our implementation as is, shown in \figref{elicit},
elicits behavioral data via binary queries
asking the user to select the optional car the she likes more than the other.

\begin{figure}[!ht]
  \centering
    \includegraphics[width=0.5\textwidth]{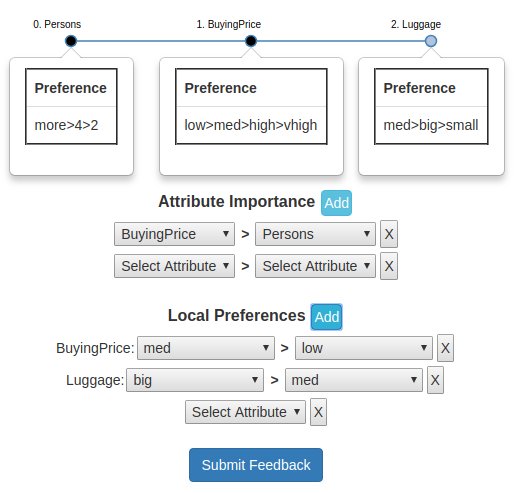}
  \caption{Visualization and feedback for UIUP trees \label{fig:feedback}}
\end{figure}

Another type of data provided by the user is the \tit{feedback data},
which are critiques based on the user's input -- visually explained model.
Clearly, feedback data are not provided by the decision maker in the initial
iteration, for no model is learned yet.
In general, feedback data are for the learning algorithm to adjust the learned
model accordingly.
In our current implementation for UIUP LP-tree models, 
the user can describe her feedback on the order of some attributes and
the order of some attributes' values.
For instance, as shown in \figref{feedback}, we see the learned UIUP tree
presented. Based on it, the user provides the feedback that BuyingPrice
should be more important than Persons.  Also, she actually prefers
medium to low on BuyingPrice, and big to medium on Luggage.

\subsection{Preprocessor and Learner}
The preprocessor takes the behavioral and feedback data and applies text mining
techniques to formalize the data to be ready for the learning module.
The learner then takes the domain description and examples and learns a model.
Currently, we implemented the greedy heuristic for learning UIUP, UICP and
CICP trees and forests of these trees \cite{conf/gcai16/liuT}.
This implementation is augmented to handle feedback data from the user,
in such a way that these data are treated as hard constraints.

\subsection{Visualizer}
\begin{figure}[!ht]
  \centering
    \includegraphics[width=0.45\textwidth]{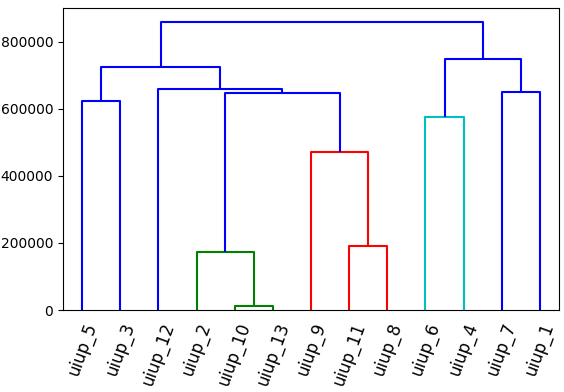}
  \caption{Clusters of UIUP trees \label{fig:clustering}}
\end{figure}

To present the learned model to the user, the visualizer draws the model and
provides annotated description of it.
Our prototype implements this module for visualizing LP-trees and LP-forests
(cf. \figref{feedback} for a UIUP tree model).
An LP-tree is drawn with expandable nodes to show or hide subtrees.
It is up to only a few levels of the tree, as deeper attributes
are less important in the model.
When a forest of trees are learned, the framework only visualizes
a very small number of representatives, 
selected by some clustering algorithm.
Our prototype applies a single-link clustering
algorithm SLINK, by Sibson \cite{sibson1973slink}, 
based on pairwise Kendall's $\tau$ distances between individual trees.
Taking as input the $\tau(T,T')$ distances between all pairs of trees, 
SLINK starts with clusters of single trees.
It merges two clusters with the minimum distance between them, where
the distance between clusters are defined as the average of distances between
all pairs of trees in them.
For example, we see the dendrogram of 13 UIUP trees in \figref{clustering},
where the y-axis are the threshold values that are the numbers of
disagreed examples between clusters of trees.
To select the representatives using this dendrogram, we simply 
use a cut-off threshold value to partition
the trees into buckets and select one model from each bucket.

%\section{Results}
%\input{results}

\section{Conclusion and Future Work}
In this paper, we presented a novel human-in-the-loop framework to help
the human user understand the decision making process of choosing from alternatives.
We focused on alternatives from combinatorial domains and qualitative preference
models that are intuitive and explainable, such as LP-trees, LP-forests, and
CP-nets.
This framework, which we call ILPref, is an interactive and iterative system.
It visualizes the learned model to the user for feedback, which is taken to improve
the model in the following iterations.
To this end, we discussed the key aspects of our prototype system 
for learning the LP models.

For future work, 
we are interested in extending our prototype to enclose more preference models,
e.g., CP-nets.
We also plan to perform a thorough user case study with human subjects
to evaluate our system, and our selected decision models.

\bibliographystyle{aaai}
\bibliography{refs}

\end{document}